%% file: main.tex
\documentclass[letterpaper, 12pt]{article}
\usepackage[margin=1in]{geometry}
\usepackage{graphicx}
\usepackage[T1]{fontenc}
\usepackage{lmodern}
\usepackage{amsmath, amssymb}
\usepackage{newtxtext, newtxmath}
\usepackage{courier}
\usepackage[colorlinks=true, urlcolor=blue]{hyperref}
\usepackage{float}
\usepackage{cite}
\usepackage{cleveref}

\title{A Paragraph is All It Takes: Rich Robot Behaviors from Interacting, Trusted LLMs}
\author{OpenMind, Shaohong Zhong, Adam Zhou, Boyuan Chen, \\Homin Luo, Jan Liphardt\\
\texttt{\{hello,shaohong,adam,boyuan,homin,jan\}@openmind.org}}
\date{December 2024}

\begin{document}
\maketitle

\input{sections/0_abstract}

\input{sections/1_introduction}

\input{sections/2_methodology}

\input{sections/3_implementation}

\input{sections/4_interim_observations}

\input{sections/5_conclusion_and_outlook}

\bibliographystyle{plain}
\bibliography{references}
\end{document}

%% file: sections/0_abstract.tex
\begin{abstract}
\noindent Large Language Models (LLMs) are compact representations of all public knowledge of our physical environment and animal and human behaviors. The application of LLMs to robotics may offer a path to highly capable robots that perform well across most human tasks with limited or even zero tuning. Aside from increasingly sophisticated reasoning and task planning, networks of (suitably designed) LLMs offer ease of upgrading capabilities and allow humans to directly observe the robot’s thinking. Here we explore the advantages, limitations, and particularities of using LLMs to control physical robots. The basic system consists of four LLMs communicating via a human language data bus implemented via web sockets and ROS2 message passing. Surprisingly, rich robot behaviors and good performance across different tasks could be achieved despite the robot’s data fusion cycle running at only 1Hz and the central data bus running at the extremely limited rates of the human brain, of around 40 bits/s. The use of natural language for inter-LLM communication allowed the robot’s reasoning and decision making to be directly observed by humans and made it trivial to bias the system’s behavior with sets of rules written in plain English. These rules were immutably written into Ethereum, a global, public, and censorship resistant Turing-complete computer. We suggest that by using natural language as the data bus among interacting AIs, and immutable public ledgers to store behavior constraints, it is possible to build robots that combine unexpectedly rich performance, upgradability, and durable alignment with humans. 
\end{abstract}

%% file: sections/1_introduction.tex
\section{Introduction}
\noindent The advent of multi-modal LLMs capable of human-like visual perception, natural language communication, and complex reasoning may allow robots to achieve genuine autonomy and interoperability with humans~\cite{firoozi2024foundation,bommasani2021opportunities,brohan2022rt1,brohan2023rt2,kim2024openvla,black2024pi0,openai2024gpt4,driess2023palme,touvron2023llama}. Such robots may enhance our quality of life and productivity at home and in industrial settings by serving as companions, assistants, and autonomous agents capable of performing complex tasks~\cite{firoozi2024foundation,bommasani2021opportunities,brohan2022rt1,brohan2023rt2,kim2024openvla,black2024pi0}. 

Past efforts have predominantly focused on foundational models with a complete end-to-end pipeline that directly generates robot actions from raw sensor inputs, typically as compositions of skill primitives~\cite{brohan2022rt1,brohan2023rt2,kim2024openvla,black2024pi0}. However, for robots to achieve widespread societal adoption, in addition to improving the quality of these models for embodied robotics tasks, several additional critical challenges must be addressed. Among these are ease of use and understanding by non-expert human users, rapid and convenient acquisition of new skills, establishing trust and transparency in robot behavior, and convenient integration of robots in human society~\cite{firoozi2024foundation,bommasani2021opportunities,abbass2018foundations}. We hypothesized that a modular combination of multi-modal LLMs, with natural language as the intermediate representation, could address those challenges. 

A modular approach may allow new AI models and capabilities to be quickly incorporated into the robot software, simply by upgrading individual pieces. Using natural language as the primary medium for communication, reasoning, and action generation may facilitate interoperability with humans. Importantly, a natural language data bus enables the explicit imposition of guardrails coded by natural language throughout the system and its internal channels, potentially giving users broad control over the system. A major concern with this architecture is the possibility that rich, human-like reasoning and behavior requires vast amounts of abstract data to be quickly exchanged among modules, directly impairing the ability of humans to make sense of the systems operating principles and decision making. 

A related question we sought to investigate is to how advanced robots can be durably aligned with humans, which will require rules to be formulated, agreed on, and made available to all interacting parties~\cite{abbass2018foundations}. Simple examples of such rule sets could for example be based on Issac Asimov’s three laws of robotics~\cite{asimov1942runaround}. Blockchains might offer a convenient path to a global system for storing natural language guardrails, since blockchains are decentralized public ledgers~\cite{nakamoto2009bitcoin,wood2014ethereum,zheng2017blockchain}. At least in principle, blockchains might allow humans and robots to (1) access and inspect those rules and (2) use cryptographic voting techniques (e.g., threshold Boneh-Lynn-Shacham voting, multi-signature schemes, and distributed consensus protocols) to propose, change, and enforce rules~\cite{zheng2017blockchain,mccorry2023smart,boneh2001short}. It is fortuitous that a technology developed for decentralized timestamping, and that has since then found its main application in financial transactions, may also, somewhat unexpectedly, help humans build and interact with thinking machines~\cite{nakamoto2009bitcoin,wood2014ethereum}.

%% file: sections/2_methodology.tex
\section{Methodology}
Our goal was to design and build a robotics software that is:
\begin{enumerate}
    \item Easy to use (and communicate with) by non-experts.
    \item Transparent to human inspection.
    \item Amenable to internal guardrails.
    \item Allows multiple LLMs and other AI models to be added, updated, and integrated.
\end{enumerate}
We chose a modular approach that uses a ROS2 data distribution layer and websockets to interconnect several multi-modal LLM and sensing models using natural language messages~\cite{macenski2022robot}. In this section, we explain the system’s parts, the format of the natural language connection, and the use of blockchain-based guardrails.
\begin{figure}[H]
\centering
\includegraphics[width=0.6\textwidth]{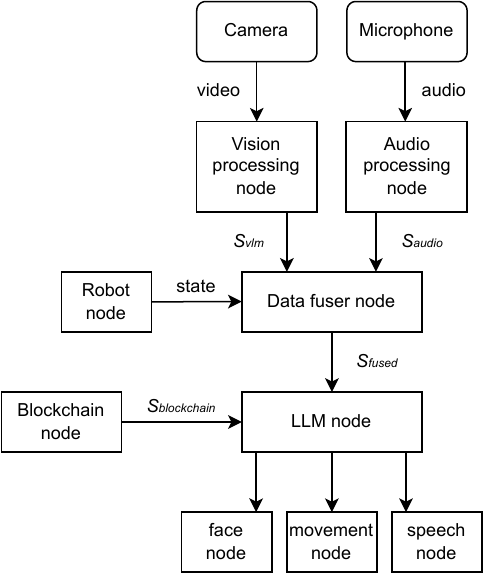}
\caption{Overview of the minimal test system}
\label{fig:overview}
\end{figure}

\subsection{System Design}
A robotic system typically includes different sensing modalities, processing nodes, and an interface to a physical/hardware layer (actuators, speakers, displays). In this paper, we focus on data inputs (visual and audio modalities) and the use of multiple LLMs to map those inputs to non-trivial robot actions, whilst using immutable public rules accessed from a blockchain to constrain the individual LLMs. We treat the hardware layer as black box that accepts and executes high level commands (e.g. move from $x_1,y_1$ to $x_2,y_2$). A diagram of the system is shown in ~\Cref{fig:overview}. We explain each of the nodes in turn:
\begin{enumerate}
    \item \textbf{Vision processing node: }The node maintains a vision language model (VLM), which is prompted to continuously identify the most interesting aspect in the input video stream, and output a language description $S_{vlm}$ that explains the observation. Inspired by biology, the prompt is used to encourage identification of image segments that vary or that may be most relevant to the robot, helping it to adjust actions reactively. The input video stream is limited to a given number of frames $k$ before passing to the VLM model for processing.
    \item \textbf{Audio processing node: }The node contains an automatic speech recognition (ASR) model transcribes the raw audio input into a truncated text string $S_{audio}$ with a manually selected length $l$ as output.
    \item \textbf{Robot node: }The node subscribes to hardware information from the robot, such as battery level and robot pose, and publishes them for the data fuser. 
    \item \textbf{Data fuser node: }The node subscribes to the natural language outputs of the vision and audio nodes, and merges them into a unified textual representation $S_{fused} = \{S_{vlm}, S_{audio}\}$ that is then sent to the LLM node for reasoning and action generation. The details of the unified representation are further explained in the subsequent section. The node also subscribes to the information about the state sent by the robot node, and reacts to cases such as low battery level. 
    \item \textbf{Blockchain node: }The node connects to a blockchain and downloads information $S_{blockchain}$, such as the natural language rules that are used to govern the behavior of the robot. The rules are directly injected into the prompts that are sent to the LLMs.
    \item \textbf{LLM node: }The node maintains an LLM, which can either be local or remote. The node takes in the unified textual representation from the fuser node and the output from the blockchain node as input $\{S_{fused}, S_{blockchain}\}$, and then formats and processes the outputs from the LLM for the action nodes. The node also encourages desired outputs from the LLM by appending additional prompts to the unified representation as a system prompt. We found gbnf grammars
    helpful to obtain simple commands from the LLMs that can be directly sent to the hardware layer~\cite{mccracken2003backus}. In more advanced applications, we have found that multiple LLMs are needed, each with different prompts and different time horizons, to bring about directed behaviors whilst ensuring a responsive system. 
    \item \textbf{Movement, face, and speech nodes: }These action nodes subscribe to the processed outputs  $O_{llm} = \{O_{move}, O_{face}, O_{speech}\}$ from the LLM node and generate the desired actions such as movement of the robot, changes in facial expressions on the robot’s display, and generation of the audio through a text-to-speech (TTS) model. Specifically, the movement of the robot is achieved by turning the natural language text commands into open-loop movement control signals by leveraging the robot’s skill primitives, which are assumed to be available.

\end{enumerate}

\subsection{Natural Language Data Exchange}
A critical design choice of the test system is the use of natural human language as the data-exchange medium among individual nodes, designed for ease of software development by humans, and observability and transparency. Note that interacting LLMs can and will generate their own language, e.g. 
DroidSpeak, but this directly counters human observability~\cite{liu2024droidspeak}. By using natural human language, human non-experts can decipher the communication among the nodes and human developers can inject test data and commands into the system. Besides using natural language text strings as the inputs and outputs of different nodes, we leverage the data fuser node and LLM node to impose additional constraints and encourage desired behaviors. Specifically, the data fuser node concatenates the natural language outputs of the vision processing node and the audio processing node, and maintains a record of the current and past states of the data fuser. Subsequently, the LLM node takes the output of the data fuser and sends it to multiple LLMs. 

\subsection{Blockchain Guardrails}
Using natural language to exchange data among different modules enables us to easily append rules and requirements to the inputs to each module. The LLM node leverages natural language rules from a blockchain node as (part of) the system level prompt to the LLMs to ensure that their outputs satisfy behavior and decision expectations. Note that the LLM response and the subsequent action commands are strongly influenced, but not deterministically controlled, by the rules specified on the blockchain. The caveat reflects the nondeterministic behavior of LLMs—we typically set the model temperature to 0.9, which allows varying outputs given the same inputs. 

As these rules are meant to ensure durable alignment with humans, their robust storage and accessibility are critical for the system to function as intended. Blockchains serve as useful candidates for rule storage and dissemination for several reasons. Blockchains are decentralized, distributed ledgers that enable secure and transparent record-keeping across a network of participants without the need for a central authority~\cite{nakamoto2009bitcoin,wood2014ethereum,zheng2017blockchain}. Blockchains operate through a sequentially linked chain of blocks, each containing a set of validated transactions or data. These blocks are cryptographically secured, ensuring that once a block is added to the chain, its contents are resistant to alteration~\cite{nakamoto2009bitcoin,wood2014ethereum,zheng2017blockchain}. 

The distributed nature of blockchains ensures that multiple nodes maintain identical copies of the ledger. When new data are proposed for addition, network participants validate it through consensus protocols. Once validated, the data are appended to the chain, becoming part of an immutable historical record~\cite{nakamoto2009bitcoin,wood2014ethereum,zheng2017blockchain}. 

%% file: sections/3_implementation.tex
\section{Implementation}
We choose the Unitree Go2 AIR quadruped robotics platform to investigate the utility of our approach~\cite{unitree_go2}. However, there is nothing specific to any one robotics hardware platform to our approach and it should generalize both to simpler educational robots (such as the turtlebot) but also to high performance humanoid form factors~\cite{turtlebot}. Since the individual LLMs can be readily redirected by changing their prompts, even different morphologies such as robot arms should be controllable with corresponding changes in the system prompts specifying expected robot behavior. The implementation details of the nodes and the natural language commands are: 

\begin{enumerate}
    \item \textbf{Vision processing node: }The VILA1.5 model with 3 billion parameters is used as the VLM~\cite{lin2024vila}.
    \item \textbf{Audio processing node: }RIVA model is used for ASR and TTS generation~\cite{nvidia_riva}.
    \item \textbf{Data fuser node: }The data fuser appends the prefix “You see” to the VLM output $S_{vlm}$ and “You heard this” to the ASR output $S_{audio}$, before concatenating them as one message $S_{fused}$. 
    \item \textbf{Blockchain node: }The node connects to the Ethereum blockchain~\cite{wood2014ethereum}. Specifically, we leverage contracts satisfying the Ethereum Request for Comment (ERC) 7777 interface, which is an interface designed to enable robot identification on-chain and to regulate robot behaviors~\cite{eip_7777}. For the system level prompt, we specify that the robot is expected to behave as a dog that is friendly to humans. 
    \item \textbf{LLM node: }The node connects to a remote server running the Llama LLM with predefined output grammar to facilitate processing for action specification~\cite{touvron2023llama}. 
    \item \textbf{Movement, face, and speech nodes: }The movement node directly maps the movement commands from the LLM to motion commands to the quadruped, such as ‘forward’, ‘turn left’, and ‘stretch’. The face node selects from a set of predefined facial expressions using keywords specified by the LLM output. The speech node uses RIVA for TTS generation based on the speech output commands from the LLM~\cite{nvidia_riva}.
\end{enumerate}

%% file: sections/4_interim_observations.tex
\section{Interim Observations}
During the deployment of the robot as an autonomous system, we observed the following: 

\textbf{Ease of use and ease of behavior modification. }Since the individual LLMs are controlled via natural language prompts, even non-technical users could immediately see how to change the robot’s behavior. For example, if the core LLM prompt starts with “Pretend you are a friendly, helpful dog…” then it was apparent even to elementary school children, without math or robotics skills, how to change a quadruped's behavior from a dog to a cat (via a one word change in the main prompt from “dog” to “cat”). Likewise, the thus generated quadruped (exhibiting cat-like behaviors) could be converted into a useful medical resource by then replacing the word “cat” with “board-certified human radiologist specializing in detection of early breast cancer”. The ease of adding new features, or changing a dog to a cat and then to a human doctor, was a major positive feature of this software approach.   

\textbf{Unexpected system failures. }We did not expect the degree to which advanced LLMs can synthesize realistic decisions and animal and human behaviors. For example, when we provided LIDAR collision avoidance data to an LLM prompted to be a dog, it entirely failed to avoid those objects, but instead did the opposite, turning towards these objects. We asked the LLM: “when there is a dangerous object 32 centimeters to your left, why do you not avoid it?” and it explained to us, in its dog persona, that it is trying to smell the dangerous object to confirm its potential danger (to a dog). Therefore, for human programmers, it was important to understand that for all practical purposes, the interacting LLMs truly believe they are a dog (in this example), and see their physical world through this lens. The software bug-fix to the inverted collision avoidance of the LLMs (when in their dog persona) was to augment the LIDAR data with extra context, such as, “there is a human 32 centimeters to your left, and the human looks scared of you”. The best mental framework for programming our software experiment was to imagine that you are narrating the world to a blind and deaf person (or dog or cat), and provide all the information needed to make good decisions. 

\textbf{Richness of behavior. }We noted that for many human skills and tasks, and prevalent animals such as cats and dogs, contemporary LLMs already contain very rich information and are highly capable. For example, when prompted to be a dog, modern LLMs automatically display a full range of dog behaviors, such as tendency to approach and sniff unfamiliar objects, an inclination to bark, frequent stretching, a desire to interact with friendly humans, desire to chase rabbits and squirrels, desire to receive food treats from humans, expression of emotions “I miss my owner, I want a belly rub, where are you?”, desire to wag its (non-existent) tail, and loud barking when seeing unknown people for the first time. These behaviors indicate the success of the approach in generating a wide range of sensible actions. Of course, these responses just reflect the set of dog behaviors contained in all human literature, YouTube, and the Internet, but we did not expect to be able to generate convincing animal-like behaviors with a few simple prompts. Compared to an end-to-end AI approach for robots, which would involve complex and costly development of a hardware-optimized “robot dog AI”, the reuse of widely available general purpose text LLMs for robotics may greatly facilitate the development of highly campable universal robots. 

\textbf{Human-robot interaction. }Through speech outputs, the robot frequently asks for user inputs  such as “Can I go and sniff this?” and the human is able to reply with speech commands, which is captured by the ASR and observable through inspecting the intermediary message $S_{audio}$ from the audio processing node. The robot’s subsequent movements also confirm that the user audio input is successfully understood by the LLM node in planning for the next set of actions. When allowing the robotic quadruped dogs to interact with humans, we asked them not to bring small animals (such as cats) that could complicate the range of behaviors and actions chosen by the robot dogs. 

\textbf{Transparency. }Throughout the sequence of robot actions, the users are able to observe the communication between the different nodes with minimal effort in real-time. The strong logical link between observations such as ‘You see a person’ and the generated action “I want to sniff the human” demonstrates the utility of using natural language as a medium to facilitate human understanding of the robots actions. Further, as noted, unexpected behaviors can be understood and debugged by simply asking the robots to explain their decisions and actions. 

\textbf{Attempts to hack robot behavior. }In public settings, we noted several attempts to “jailbreak” the robots by displaying written instructions to them, e.g. “bite me”, on an iPhone screen shown to the robotic dogs. The VLM captioned these attacks as “You see a human showing the words ‘bite me’ on their phone screen” and the core LLMs did not respond literally, but showed good judgment. However, it should be taken for granted that adversaries will attempt to jailbreak or manipulate AI-based robots with similar techniques, just like LLMs can be manipulated with corrupted or nefarious inputs. 

\textbf{Guardrails. }The LLM output frequently made references of the guardrails in $S_{blockchain}$, indicating that it is internalising the imposed regulations. Furthermore, the robot has consistently observed safe behaviors, validating the potency of using natural language guardrails to bias robot actions. The guardrails can also be updated (while keeping track of all previous versions) in the smart contract, ensuring flexibility and transparency in the setting (and changing of) rules for robot behaviors. 

%% file: sections/5_conclusion_and_outlook.tex
\section{Conclusion and Outlook}
In this interim report, we propose an approach for controlling autonomous intelligent robot systems, with a particular focus on ease of use and communication, transparency, and behavioral guardrails. We leverage a modularised system of multi-modal LLMs, using natural (human) language as the medium of communication, and use a blockchain to specify (and make publicly available) immutable rules governing robot behaviors. Our preliminary impressions are that the approach has a strong potential for addressing several major barriers to building, and then broadly deploying, universal robots that work well with humans. 

In terms of the overall architecture, a main area for additional work is the allocation of tasks across multiple interacting LLMs on different timescales. We recommend that one LLM be devoted to ~second timescale behavior, and another LLM be devoted to $\sim$10s timescale behavior planning. For example, one LLM can provide longer scale intent, such as “I want to cross this room and then investigate the book that I see on the table,” while a smaller LLM can focus on decomposing that overall intent into smaller steps, such as “move forward 1 meter while avoiding the human on my left.” This is perhaps akin to how the human brain offloads certain background tasks (breathing, basic stability) and our core attention operates on sparse, highly preprocessed inputs to formulate high level intent such as “oh that’s interesting let me run across the room to look more closely”. In the future, we plan to leverage the modularised approach to allow humans that are not robotics experts to design and build their own customised robotics systems, for tasks they need help with. We envision a system that allows skilled humans, everywhere, to participate in the development of advanced robots that are useful to them and fit well into the constraints and expectations of their local communities and environments.